\documentclass[sigconf]{acmart}

\usepackage{multirow}
\usepackage{cite}
\usepackage{algorithm} 
\usepackage{algorithmic}
\usepackage{graphicx}
\usepackage{xcolor}
\usepackage{subcaption}
\usepackage{balance}
\AtBeginDocument{%
  }

\copyrightyear{2023} \acmYear{2023} \setcopyright{acmlicensed}\acmConference[CIKM '23]{Proceedings of the 32nd ACM International Conference on Information and Knowledge Management}{October 21--25, 2023}{Birmingham, United Kingdom} \acmBooktitle{Proceedings of the 32nd ACM International Conference on Information and Knowledge Management (CIKM '23), October 21--25, 2023, Birmingham, United Kingdom} \acmPrice{15.00} \acmDOI{10.1145/3583780.3614956} \acmISBN{979-8-4007-0124-5/23/10}

\begin{document}

\title{MadSGM: Multivariate Anomaly Detection with Score-based Generative Models}

\author{Haksoo Lim}
\authornote{Both authors contributed equally to this research.}
\affiliation{%
  \institution{Yonsei University}
  \streetaddress{50, Yonsei-ro}
  \city{Seoul}
  \country{Republic of Korea}
  \postcode{03722}
}
\email{limhaksoo96@yonsei.ac.kr}
\orcid{0000-0003-3182-5948}

\author{Sewon Park}
\authornotemark[1]
\affiliation{%
  \institution{Samsung SDS}
  \streetaddress{125, Olympic-ro 35-gil}
  \city{Seoul}
  \country{Republic of Korea}
  \postcode{05510}
}
\email{sw0413.park@samsung.com}
\orcid{0000-0002-6811-4632}

\author{Minjung Kim}
\authornotemark[1]
\affiliation{%
  \institution{Samsung SDS}
  \streetaddress{125, Olympic-ro 35-gil}
  \city{Seoul}
  \country{Republic of Korea}
  \postcode{05510}
}
\email{mj100.kim@samsung.com}
\orcid{0009-0002-3652-8475}

\author{Jaehoon Lee}
\authornote{This work was done when he was at Yonsei university.}
\affiliation{%
  \institution{LG AI Research}
  \streetaddress{30, Magokjungang 10-ro}
  \city{Seoul}
  \country{Republic of Korea}
  \postcode{07796}
}
\email{jaehoon.lee@lgresearch.ai}
\orcid{0009-0002-0744-6593}

\author{Seonkyu Lim}
\affiliation{%
  \institution{Yonsei University}
  \streetaddress{50, Yonsei-ro}
  \city{Seoul}
  \country{Republic of Korea}
  \postcode{03722}
}
\email{seonkyu@yonsei.ac.kr}
\orcid{0000-0003-4904-7144}

\author{Noseong Park}
\affiliation{%
  \institution{Yonsei University}
  \streetaddress{50, Yonsei-ro}
  \city{Seoul}
  \country{Republic of Korea}
  \postcode{03722}
}
\email{noseong@yonsei.ac.kr}
\orcid{0000-0002-1268-840X}

\renewcommand{\shortauthors}{Lim et al.}

\begin{abstract}
The time-series anomaly detection is one of the most fundamental tasks for time-series. Unlike the time-series forecasting and classification, the time-series anomaly detection typically requires unsupervised (or self-supervised) training since collecting and labeling anomalous observations are difficult. In addition, most existing methods resort to limited forms of anomaly measurements and therefore, it is not clear whether they are optimal in all circumstances. To this end, we present a multivariate time-series anomaly detector based on score-based generative models, called MadSGM, which considers the broadest ever set of anomaly measurement factors: i) reconstruction-based, ii) density-based, and iii) gradient-based anomaly measurements. We also design a conditional score network and its denoising score matching loss for the time-series anomaly detection. Experiments on five real-world benchmark datasets illustrate that MadSGM achieves the most robust and accurate predictions.
\end{abstract}




\received{20 February 2007}
\received[revised]{12 March 2009}
\received[accepted]{5 June 2009}

\maketitle
 
\section{Introduction}
\label{introduction}


Time-series-based applications are abundant in our daily life. For instance, traffic condition monitoring systems~\citep{wu2018traffic}, and industrial/scientific remote sensing systems~\citep{dong2019remote} are their representative examples. One common task in those applications is to detect anomalous observations, which may cause severe damage to our society. However, it is hard to label anomalous observations and perform supervised training. Therefore, most of the well-known time-series anomaly detection algorithms are unsupervised (or self-supervised). These existing methods are known to be successful for many time-series datasets. Nevertheless, one limitation of most existing detection methods is that they resort to a single type of anomaly measurement, such as reconstruction-based or density-based anomaly measurement (see Table~\ref{tab:anoscore}). A single type of anomaly measurement may not work well in real-world time-series data due to its complicated characteristics. For example, when normal points are similar to abnormal ones in a feature space, reconstruction-based models frequently fail. Moreover, it is difficult for density-based models to discern between normal points in a low-density region and abnormal points whose probabilities are naturally low.


In order to overcome the limitations, we consider simultaneously three anomaly measurement types to detect as many anomalies as possible: i) reconstruction-based, ii) density-based, iii) gradient\footnote{The gradient in our context means the gradient of the log-density of data, which is called as \emph{score}. The term `score' of score-based generative models also means it. One may consider that the anomaly detection based on the gradient, therefore, falls into the category of the density-based type. However, the gradient-based anomaly detection yields prediction outcomes distinctive from the density-based methods.}-based anomaly measurements. 
To compute the three anomaly measurements in a robust manner, we apply score-based generative models (SGMs)~\citep{song2021SDE, kim2022sos}
to our task. Recent work has demonstrated that SGMs possess great strengths in reproducing high-quality samples and obtaining exact probability densities via score functions, i.e., the gradient of the log-density of samples. We design an SGM-based method for our time-series task since SGMs have been typically studied for images and there exist a few papers for the time-series forecasting only. All three anomaly measurements can be acquired naturally through the training and sampling procedures of our proposed SGM method. To be more specific, we i) design our own conditional score network and ii) redesign the denoising score matching loss~\citep{lim2023tsgm} for our time-series-based task, which is specialized for capturing subtle points between normal observations and anomalies. Our designed conditional SGM method, which consists of a conditional score network and its denoising score matching training, is basically autoregressive since it learns $\nabla_{\textbf{x}_{t-\omega:t}^l}{\text{log}p(\textbf{x}_{t-\omega:t}^l|\textbf{x}_{t-\omega:t-1}^0)}$ at the diffusion step $l$ --- in other words, our SGM model is able to sample $\textbf{x}_{t-\omega:t}^0$ given $\textbf{x}_{t-\omega:t-1}^0$.


However, there exists one subtle point in our autoregressive approach. Once an anomalous observation is fed into our method, all following observations can be predicted as anomalies in the worst case --- in other words, the prior anomalous observation influences its following normal observations, which is not preferred since our task is to pinpoint a narrow temporal window with anomalies. Thus, we need to prevent the propagation of anomaly decisions and do it via a purification step, which is the process of calibrating anomaly measurement (see Sec.~\ref{Sec:purification}). The purification step consists of two operations: noising and denoising. First, we add noises to the prior observations and then denoise to derive purified observations. At the end, we use them as a condition for sampling the next observation during our autoregressive processing. We found that this approach significantly stabilizes the overall processing and therefore, we can extract the reconstruction-based, density-based, and gradient-based anomaly measurements in a robust manner --- we call them as \emph{calibrated anomaly measurements} in order to distinguish them from na\"ive ones (cf. Sec.~\ref{Sec:anomalyscore} vs. Sec.~\ref{Sec:purification}).



We conduct experiments on five benchmark datasets with nine baselines. To assess the detection performance, we mainly use the F1-score with the PA\%K strategy as our main evaluation metric, which is known to be more appropriate for the time-series anomaly detection. We also consider the widely-used point adjustment approach. Our model shows the best detection performance in terms of the two evaluation metrics on almost all datasets. Our contributions can be summarized as follows:



\begin{enumerate}
    \item To our best knowledge, we present a time-series anomaly detection method based on SGMs for the first time and consider the broadest ever set of anomaly measurements: i) the reconstruction-based, ii) the density-based, and iii) the gradient-based ones.
    \item We train our conditional score network by using our proposed denoising score matching loss, which is specially designed for time-series anomaly detection. We also adopt the purification strategy to achieve the fairness in the decision process by preventing the propagation of anomaly decisions.
    \item We conduct comprehensive experiments on five benchmark datasets for the time-series anomaly detection. Our results illustrate that MadSGM has better and more robust performance than baselines.
\end{enumerate}

\begin{figure*}[t]
\centering
\includegraphics[width=0.9\textwidth]{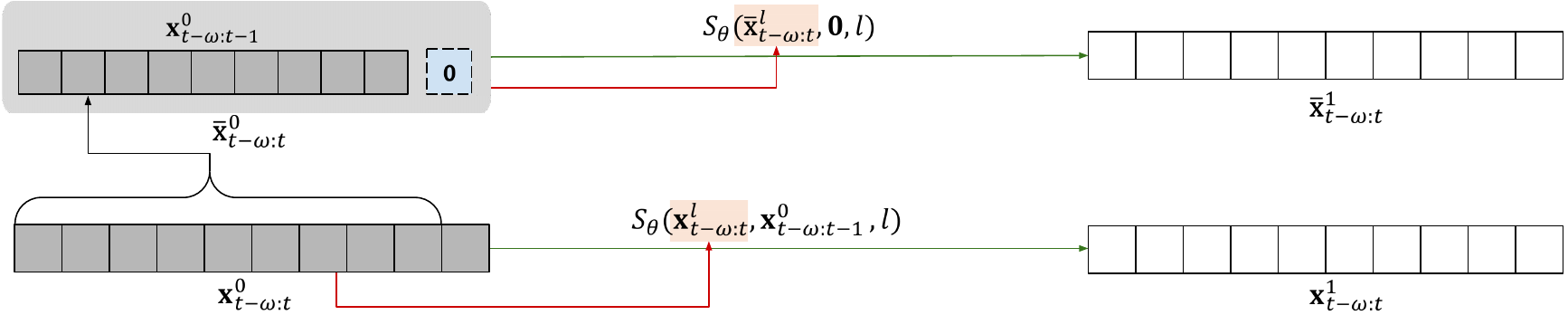}
\caption{The overall framework of training process. Note that each score network learns $S_{\boldsymbol{\theta}}(\textbf{x}_{t-\omega:t}^l, \textbf{x}_{t-\omega:t-1}^0, l) \approx \nabla_{\textbf{x}_{t-\omega:t}^l}{\text{log}p(\textbf{x}_{t-\omega:t}^l|\textbf{x}_{t-\omega:t-1}^0)}$ and $S_{\boldsymbol\theta}(\bar{\textbf{x}}_{t-\omega:t}^l,\textbf{0},l) \approx {\nabla}_{\bar{\textbf{x}}_{t-\omega:t}^l}\text{log}p(\bar{\textbf{x}}_{t-\omega:t}^l)$, respectively (cf. Section 3.3).}
\vspace{-0.3cm}
\label{fig:training}
\end{figure*}

\section{Related Works}

\subsection{Anomaly Detection}

Anomaly detection is the process to find rare observations deviating from a normal pattern distribution.
However, because abnormal observations are scarce~\citep{adsurvey} and it is hard to collect and label abnormal observations, we cannot easily apply supervised learning to anomaly detection. Therefore, an unsupervised setting is common in an anomaly detection task. There are several methods for this task: i) reconstruction-based, ii) density-based, and iii) boundary-based methods. Each method is characterized by the assumption about the characteristics of anomalies.




The reconstruction-based methods assume that anomalous samples can't be accurately reconstructed by their trained models. Their anomaly measurements are based on the reconstruction error. LSTM-VAE~\citep{DBLP:journals/corr/abs-1711-00614} utilizes LSTMs in variational autoencoder (VAE) to take into account the temporal dependency
of time-series data. OmniAnomaly~\citep{10.1145/3292500.3330672} adds a stochastic module in the LSTM-VAE to capture stochastic properties in time-series.  MAD-GAN~\citep{Li2019MADGANMA} and TAnoGAN~\citep{bashar2020tanogan} 
use a GAN architecture composed of a discriminator and a generator with LSTM layers. It detects anomalies using both discrimination and reconstruction losses. USAD~\citep{usad} employs two autoencoders and trains them with adversarial loss to isolate anomalous samples and provide fast training.  MSCRED~\citep{zhang2019deep} constructs attention-based ConvLSTM networks for temporal modeling and a convolutional autoencoder to compress and reconstruct the inter-sensor (time-series) correlation
patterns. For AT~\citep{xu2022anomaly}, they introduce the series-association from Transformers~\citep{NIPS2017_3f5ee243} and Gaussian prior-association to differentiate between normal and abnormal patterns. TadGAN~\citep{geiger2020tadgan} not only uses reconstruction error like other GAN-based methods, but also devises other measurement by using discriminator, which is called `Critic'. 
TadGAN mainly focuses on univariate time-series, but it can be generalized into multivariate time-series.

Density-based methods estimate the distribution of normal data and can compute probabilities of normal and abnormal points. The assumption of these methods is that in the estimated distribution, the probability of anomalous observations is lower than that of normal ones. LOF~\citep{10.1145/335191.335388} is a traditional method calculating local density for outlier determination.  DAGMM~\citep{zong2018deep} combines an autoencoder for dimension reduction with a finite gaussian mixture model for density estimation of latent variables. Adaptive-KD~\citep{zhang2018adaptive} estimates local densities using an adaptive kernel density estimation approach in nonlinear systems.

\begin{table}[t]
  \caption{   
    Comparison among various methods, including MadSGM, in terms of the detection method. We point out that none of the baselines consider multiple criterion.
    }
\vspace{-0.3cm}
 \setlength{\tabcolsep}{3pt}
\begin{center}  
   \begin{tabular}{c|ccc}
    \toprule
    & Reconst.-based & Density-based & Gradient-based\\
    \hline
    LSTM-VAE & \checkmark & $\times$ & $\times$\\
    MAD-GAN & \checkmark & $\times$ & $\times$\\
    USAD  & \checkmark & $\times$ & $\times$\\
    AT 	 & \checkmark & $\times$ & $\times$\\
    DAGMM & $\times$ & \checkmark & $\times$\\ 
    \hline
   MadSGM (Ours)  & \checkmark & \checkmark & \checkmark \\
    \bottomrule 
    \end{tabular}
    \vspace{-0.5cm}
  \end{center}
  \label{tab:anoscore}
\end{table}

As for the boundary-based methods, it is supposed that in a good representation space, there is a boundary that distinguishes abnormal points from normal ones. OCSVM~\citep{tax2004support} maps the training data into the feature space via kernel functions and finds an optimal hyperplane of maximal margin that separates the normal data from the origin. DeepSVDD~\citep{ruff2018deep} constructs neural networks to find a hypersphere of minimum volume that includes the normal data on the latent space. THOC~\citep{shen2020timeseries} extracts multi-scale temporal features by using a multi-layer dilated RNN and builds a hierarchical structure with multiple hyperspheres for each resolution.

Usually, extant studies are based on only one type of method. However, it is not desirable to stick to a single one. This is because the complex nature of real-world time-series data makes it hard for only one of the methods to be sufficient. Therefore, we devise our model which can utilize multiple methods at the same time. We show the effectiveness of combining multiple methods in Section~\ref{sec:abl_anomaly_score}.





\subsection{Score-based Generative Models}

SGMs~\citep{song2021SDE} diffuse a data distribution $p_0(\textbf{x})$ to a noise distribution $\pi(\textbf{x})$ with an It\^o stochastic differential equation (SDE):
\begin{align}
    d\textbf{x}=\textbf{f}(\textbf{x},l)dl+g(l)d\textbf{w}, \qquad l \in [0,1], \label{Forward}
\end{align}
where $\textbf{f}(\cdot,l):\mathbb{R}^n \rightarrow \mathbb{R}^n$ and $g:[0,1] \rightarrow \mathbb{R}$ denote drift and diffusion coefficients of $\textbf{x}^l$, respectively, and ${\textbf{w} \in \mathbb{R}^n}$ is a Brownian motion. There are several types of SDE such as variance exploding (VE), variance preserving (VP), and subVP, depending on the definition of coefficients $\textbf{f}$ and $g$ as in ~\citet{song2021SDE}. A diffusion process $\{\textbf{x}^l\}_{l\in[0,1]}$ can be derived by solving the SDE~(\ref{Forward}). By sufficiently perturbing the data $\textbf{x}^0$ using SDE, the distribution of $\textbf{x}^1$ at the end step can be approximated by the noise distribution. 

The reverse SDE for generating samples $\textbf{x}^0$ from noisy samples $\textbf{x}^1$ is as follows:
\begin{align*}
    d\textbf{x}=\left[\textbf{f}(\textbf{x},l)-g^2(l)\nabla_{\textbf{x}}{\text{log}p_l(\textbf{x})}\right]dl+g(l)d\bar{\textbf{w}}, \quad l \in [0,1],
\end{align*}
where $\nabla_{\textbf{x}}{\text{log}p_l(\textbf{x})}$ is the score function of $\textbf{x}^l$, 
$\bar{\textbf{w}}$ is a Brownian motion in the reverse time direction and $dl$ is a negative time step. In the reverse SDE, the unknown score function $\nabla_{\textbf{x}}{\text{log}p_l(\textbf{x})}$ can be estimated as a score network $S_{\boldsymbol{\theta}}(\textbf{x},l)$ using the denoising score matching~\citep{vincent2011matching, song2021SDE}. The loss function to train the score network is given by 
\begin{align*}
L(\boldsymbol{\theta})=\mathbb{E}_{l}\left\{\lambda(l)\mathbb{E}_{\textbf{x}^0}\mathbb{E}_{\textbf{x}^l|\textbf{x}^0}\left[\left\|S_{\boldsymbol{\theta}}(\textbf{x}^l,l)-{\nabla}_{\textbf{x}^l}\text{log}p(\textbf{x}^l|\textbf{x}^0)\right\|_2^2\right]\right\},
\end{align*}
where $\lambda(l) > 0$ is a weighting function and $p(\textbf{x}^l|\textbf{x}^0)$ denotes a transition kernel. Note that the transition kernel is a Gaussian distribution when the drift coefficient $\textbf{f}(\cdot,l)$ is affine as in~\citep{sarkka2019applied}.

There are two numerical approaches to solve the reverse SDE for sampling: the \textit{predictor-corrector} and using well-known ODE \textit{solvers} on probability flow ODE~\citep{DBLP:journals/corr/abs-1806-07366}. 
The ODE \textit{solver} can be used to the following ordinary differential equation (ODE) which has the same probability distribution of $\textbf{x}^l$ as that of the SDE (Eq.~\eqref{Forward}):
\begin{align*}
    d\textbf{x}=\left[\textbf{f}(\textbf{x},l)-\frac{1}{2}g^2(l)\nabla_{\textbf{x}}{\text{log}p_l(\textbf{x})}\right]dl,
\end{align*}
where $\nabla_{\textbf{x}}{\text{log}p_l(\textbf{x})}$ can be replaced by $S_{\boldsymbol{\theta}}(\textbf{x}^l,l)$. \citep{DBLP:journals/corr/abs-1806-07366} proved that one can compute the exact log-likelihood of $\textbf{x}^0$ from the formula:
\begin{align}
    \log p(\textbf{x}^0)=\log p(\textbf{x}^1)+\int_0^1 \nabla_{\textbf{x}^l} \cdot \tilde{\textbf{f}}(\textbf{x}^l,l)dl, \label{eq:pflow}
\end{align}
where $\tilde{\textbf{f}}(\textbf{x}^l, l)=\textbf{f}(\textbf{x}^l, l)-\frac{1}{2}g^2(l)\nabla_{\textbf{x}^l}{\text{log}p(\textbf{x}^l)}$. The Hutchinson’s trace estimator enables the unbiased linear-time estimation of $\nabla_{\textbf{x}^l} \cdot \tilde{\textbf{f}}(\textbf{x}^l,l)$~\citep{ffjord}. As such, we can use both the generated samples and the log-likelihood computed from the probability flow ODE to define the proposed anomaly measurement described in Section~\ref{proposed}.

\subsection{Adversarial Purification} \label{sec:ap}
Deep neural networks in the image domain are known to have a high vulnerability to adversarial attacks using the adversarially perturbed images to cause misclassification. As a defense strategy against such attacks, there is \textit{adversarial purification} that purifies perturbed images into clean images. Many existing purification methods focused on deep generative models. Defense-GAN~\citep{samangouei2018defensegan} reduces the effect of the adversarial perturbation using a WGAN-based method.
~\citet{yoon2021purification} utilizes an energy-based model trained using denoising score matching. DiffPure~\citep{nie2022purification} removes noises from attacked images via the forward and reverse SDEs in SGM~\citep{song2021SDE}, whose main intuition is that gradually reducing noises in the reverse process is similar to the role of the purification model. We further enhance our method by adopting this adversarial purification idea.

\section{Proposed Method}\label{proposed}


In this section, we describe the proposed anomaly detection method in detail. The key points in our method are that i) we use a conditional score network to preserve the temporal dependencies on time-series and introduce a denoising score matching loss function to train the conditional score network for the purpose of anomaly detection, ii) 
we define three types of anomaly measurements: a) reconstruction-based, b) probability-based, c) gradient-based anomaly measurements, and iii) we propose a calibrating strategy based on a purification method for the conditional input of our conditional score network.


\subsection{Problem Statement}
Let $\mathcal{T}=\{\textbf{x}_1,\cdots,\textbf{x}_T\}$, where $\textbf{x}_t\in \mathbb{R}^m$ is an observation at time $t$, be a multivariate time-series sequence, and $\textbf{x}_{t-\omega:t}$ $=\{\textbf{x}_{t-\omega},\cdots,\textbf{x}_t\}$ be a window of length $\omega+1$. Thus, $\mathcal{T}$ can be divided into $T-\omega$ sliding windows, i.e., $\{\mathcal{T}_j\}_{j=1}^{N}$, where $N = T-\omega$. We consider the challenging environment that all observations in our training data $\{\mathcal{T}_j\}_{j=1}^{N}$ are unlabeled, i.e., unsupervised training. When training our model and computing the anomaly measurement, our prediction granularity is a window $\textbf{x}_{t-\omega:t}$ $=\{\textbf{x}_{t-\omega},\cdots,\textbf{x}_t\}$. In other words, our task is to detect whether each window has anomalies or not.

\subsection{Score Network for Time-series Anomaly Detection}

For non-sequential data such as images~\citep{SCHLEGL201930} and tables~\citep{shenkar2022ADtable}, anomaly detection methods can independently determine whether each sample is abnormal or not, whereas in the time-series domain there exist temporal dependencies among observations, requiring a different approach utilizing them.
Therefore, to capture the conditional data distribution $p(\textbf{x}_{t-\omega:t}|\textbf{x}_{t-\omega:t-1})$, the score network $S_{\boldsymbol{\theta}}(\cdot, \cdot, \cdot)$ must take 3 inputs: a diffusion step $l$, a diffused sample $\textbf{x}_{t-\omega:t}^l$, and a condition $\textbf{x}_{t-\omega:t-1}$. The conditional score network $S_{\boldsymbol{\theta}}(\textbf{x}_{t-\omega:t}^l, \textbf{x}_{t-\omega:t-1}, l)$ estimates the gradient of the conditional log probability $\nabla_{\textbf{x}_{t-\omega:t}^l}{\text{log}p(\textbf{x}_{t-\omega:t}^l|\textbf{x}_{t-\omega:t-1})}$, called as conditional score function.

In order to design our conditional score network, we modify the reputed U-net architecture~\citep{ronneberger2015unet} to capture temporal dependencies better. From the U-net architecture, we replace its 2-dimensional convolutional layers with 1-dimensional ones and follow the miscellaneous structures in~\citep{song2019smld,song2021SDE}. After concatenating the diffused sample and the condition, we feed it into our conditional score network. We refer the readers to Section~\ref{Sec:hyperparameters} for the detailed architecture and its hyperparameters.




\subsection{Training and Sampling Methods} \label{Sec:trainingsampling}
We redesign the denoising score matching~\citep{vincent2011matching, song2021SDE, lim2023tsgm} for our sake, and our proposed conditional score network learns the time-series patterns from the training data. The parameters of score network can be trained by minimizing 
\begin{equation}
 L_{score}(t) = \mathbb{E}_{l}\mathbb{E}_{\textbf{x}_{t-\omega:t}}\left[\lambda(l)(L_1(t)+L_2(t))\right],
\end{equation}where
\begin{small}
\begin{align*}
&L_1(t) = \mathbb{E}_{{\textbf{\textsc{x}}}_{t-\omega:t}^l}\left[\left\|S_{\boldsymbol\theta}({\textbf{\textsc{x}}}_{t-\omega:t}^l,\textbf{\textsc{x}}_{t-\omega:t-1}^0,l)-{\nabla}_{{\textbf{\textsc{x}}}_{t-\omega:t}^l}\text{log}p({\textbf{\textsc{x}}}_{t-\omega:t}^l|\textbf{\textsc{x}}_{t-\omega:t}^0)\right\|_2^2\right],\\
&L_2(t) = \mathbb{E}_{\bar{\textbf{x}}_{t-\omega:t}^l}\left[\left\|S_{\boldsymbol\theta}(\bar{\textbf{x}}_{t-\omega:t}^l,\textbf{0},l)-{\nabla}_{\bar{\textbf{x}}_{t-\omega:t}^l}\text{log}p(\bar{\textbf{x}}_{t-\omega:t}^l|\bar{\textbf{x}}_{t-\omega:t}^0)\right\|_2^2\right].
\end{align*}
\end{small}
Here, $\lambda(l) > 0$ is a weighting function as in~\citet{song2021SDE}. 
Note that our conditional score network is trained for both $L_1(t)$ and $L_2(t)$. As such, for $L_2(t)$, so to use $\textbf{x}_{t-\omega:t-1}$ as the input of the first part of the $S_{\boldsymbol{\theta}}(\cdot,\cdot,\cdot)$, we concatenate $\textbf{x}_{t-\omega:t-1}$ and $0$ to match its dimension with $\textbf{x}_{t-\omega:t}^l$ of $L_1(t)$. We denote the concatenated data as $\bar{\textbf{x}}_{t-\omega:t}$. \citet{lim2023tsgm} proved that the denoising conditional score matching loss $L_1(t)$ is equivalent to the explicit conditional score matching loss,
\begin{small}
\begin{align*}
\mathbb{E}_{{\textbf{\textsc{x}}}_{t-\omega:t}^l}\left[\left\|S_{\boldsymbol\theta}({\textbf{\textsc{x}}}_{t-\omega:t}^l,\textbf{\textsc{x}}_{t-\omega:t-1}^0,l)-{\nabla}_{{\textbf{\textsc{x}}}_{t-\omega:t}^l}\text{log}p({\textbf{\textsc{x}}}_{t-\omega:t}^l|\textbf{\textsc{x}}_{t-\omega:t-1}^0)\right\|_2^2\right].
\end{align*}
\end{small}
\noindent Therefore, the minimization of the loss function $L_1(t)$ leads to $S_{\boldsymbol{\theta}}(\textbf{x}_{t-\omega:t}^l, \textbf{x}_{t-\omega:t-1}^0, l) \approx \nabla_{\textbf{x}_{t-\omega:t}^l}{\text{log}p(\textbf{x}_{t-\omega:t}^l|\textbf{x}_{t-\omega:t-1}^0)}$. 

In addition, because the score function in $L_2(t)$ takes zero vectors as the condition, it can be regarded as a na\"ive score matching which doesn't require any condition values. Thus, we can think $S_{\boldsymbol\theta}(\bar{\textbf{x}}_{t-\omega:t}^l,\textbf{0},l) \approx {\nabla}_{\bar{\textbf{x}}_{t-\omega:t}^l}\text{log}p(\bar{\textbf{x}}_{t-\omega:t}^l)$. We point out that $L_2(t)$ is the na\"ive denoising score matching loss as in~\citet{song2021SDE}, whereas $L_1(t)$ is the conditional score matching loss devised by~\citet{lim2023tsgm}. The conditional score function learned by $L_1(t)$ and $L_2(t)$ has different roles in anomaly detection. The conditional score function from $L_1(t)$ is used to calculate the anomaly measurement (Section~\ref{Sec:anomalyscore}) and that from $L_2(t)$ is for purification (Section~\ref{Sec:purification}) which is to adjust the conditional values of the score network in anomaly detection. The entire training process of our proposed model is shown in Figure~\ref{fig:training}.


After training conditional score network, to generate samples from a given noisy vector $\textbf{z}\sim N(\textbf{0}, \textbf{I})$ and previous data, we solve the following reverse SDE or probability flow ODE by using the \textit{predictor-corrector} or well-known ODE \textit{solver}~\citep{DBLP:journals/corr/abs-1806-07366}, respectively: 
\begin{small}
\begin{align*}
d\textbf{x}_{t-\omega:t}^l=\left[\textbf{f}(\textbf{x}_{t-\omega:t}^l,l)-g^2(l)\nabla_{\textbf{x}_{t-\omega:t}^l}{\text{log}p(\textbf{x}_{t-\omega:t}^l|\textbf{x}_{t-\omega:t-1}^0)}\right]dl+g(l)d\bar{\textbf{w}},
\end{align*}
\end{small}
\vspace{-1.0em}

\begin{small}
\begin{align*}
d\textbf{x}_{t-\omega:t}^l=\left[\textbf{f}(\textbf{x}_{t-\omega:t}^l,l)-{1 \over 2}g^2(l)\nabla_{\textbf{x}_{t-\omega:t}^l}{\text{log}p(\textbf{x}_{t-\omega:t}^l|\textbf{x}_{t-\omega:t-1}^0)}\right]dl.
\end{align*}
\end{small}

It is known that the probability flow ODE is faster but generates poorer samples than that of the reverse SDE. Up to our works, the probability flow ODE yields similar results with the reverse SDE, but it is averaged 5 times faster (see Table~\ref{tab:nfe}), so we only use the probability flow ODE for our entire experiments.
Furthermore, by the instantaneous change of variable theorem~\citep{DBLP:journals/corr/abs-1806-07366}, we can also exactly compute the conditional log-likelihood ${\text{log}p(\textbf{x}_{t}|\textbf{x}_{t-\omega:t-1})}$.


\begin{table}[h]
  \caption{   
    Comparison of the average number of function evaluations (NFE) between the probability flow ODE and the reverse SDE.
    }
\vspace{-0.2cm}
 \setlength{\tabcolsep}{3pt}
\begin{center}  
   \begin{tabular}{c|ccccc}
    \toprule
    & \multicolumn{5}{c}{NFE} \\
    \cmidrule(lr){2-6}
    & SWaT & SMAP & MSL & PSM & SMD\\
    \hline
    reverse SDE & \multicolumn{5}{c}{2000}\\
    probability flow ODE & 310.4 & 392.1 & 712.5 & 365.0 & 388.1\\
    \hline
  ratio(upper/lower) & 6.44 & 5.10 & 2.81 & 5.48 & 5.15 \\
    \bottomrule 
    \end{tabular}
    \vspace{-0.4cm}
  \end{center}
  \label{tab:nfe}
\end{table}

\begin{figure}[t]
\centering
\includegraphics[width=0.4\textwidth]{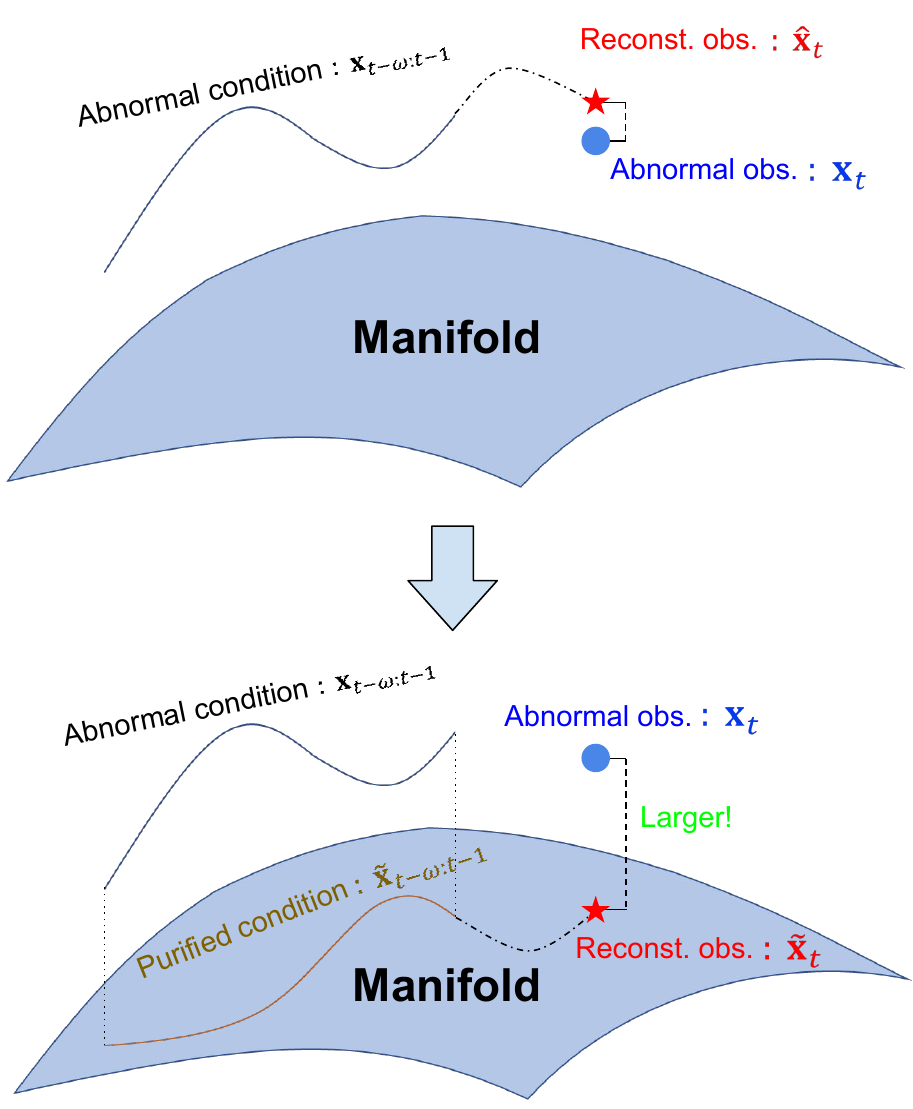} 
\caption{Comparison between the anomaly measurement without and with the purification process. The manifold represents a space where normal data distribution lies. (upper) Given the abnormal condition $\textbf{x}_{t-\omega:t-1}$, the reconstructed observation $\hat{\textbf{x}}_{t}$ will be close to the abnormal observation $\textbf{x}_{t}$. (lower) Given the purified condition, the reconstructed observation will be far from the abnormal observation.}
\label{fig:purification}
\vspace{-0.5cm}
\end{figure}

\begin{figure*}[t]
\centering
\includegraphics[width=0.9\textwidth]{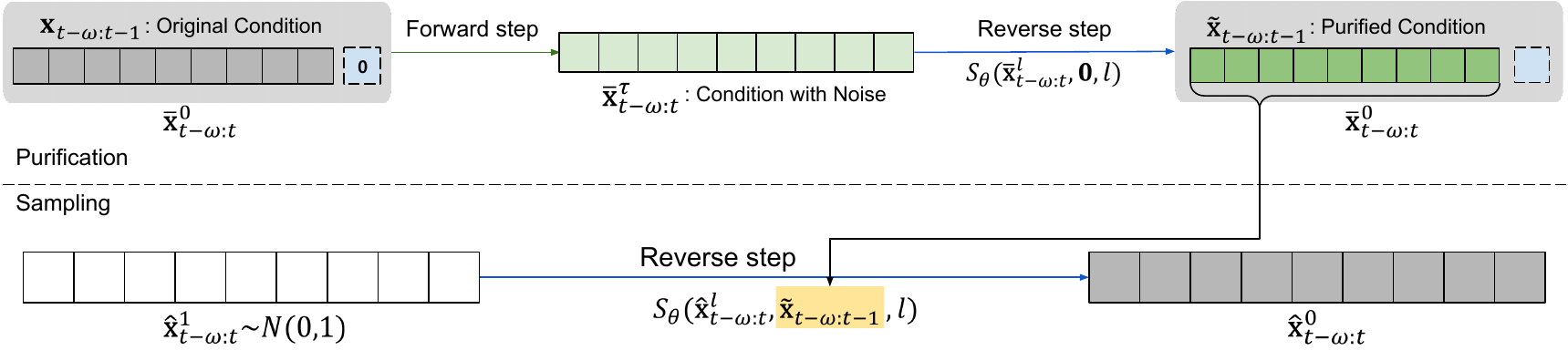} 
\vspace{-0.3cm}
\caption{Detailed process of the purification and sampling process in Figure~\ref{fig:purification}}
\vspace{-0.cm}
\label{fig:anomaly}
\end{figure*}

\subsection{Na\"ive Anomaly measurement Definitions} 
\label{Sec:anomalyscore}
Anomalies are unusual observations that deviate from normal behaviors. We define SGM-based anomaly measurements using a conditional score network that learns normal patterns from training data. In particular, to detect tricky anomalies, we use the advantages of SGM, which generates high-quality samples and provides an estimate of the gradient of log probability. We determine whether a sample for time step $t>T$ is normal or abnormal using the previous observations. The proposed anomaly measurement consists of i) reconstruction-based measurement, $A_{recon}$, ii) probability-based measurement, $A_{prob}$, and iii) gradient-based measurement, $A_{grad}$, which are explained from following subsection.

\subsubsection{Reconstruction-based Measurement} 
We generate $\hat{\textbf{x}}_{t-\omega:t}$ with trained SGMs and extract the expected value $\hat{\textbf{x}}_{t}$ at time $t$ from the normal temporal trend. We define the following reconstruction-based anomaly measurement as the difference between the observed value $\textbf{x}_t$ and the reconstructed value $\hat{\textbf{x}}_t$:
\begin{align*}
 A_{recon}(t)=\left\|\hat{\textbf{x}}_t-\textbf{x}_t\right\|_2^2.
\end{align*}

\subsubsection{Probability-based Measurement} 
In the sampling procedure of SGMs, the probability flow ODE computes the log conditional probability ${\text{log}p(\textbf{x}_{t}|\textbf{x}_{t-\omega:t-1})}$ with Eq.~\eqref{eq:pflow} that captures temporal dependencies of normal points. Due to the sparsity of anomalous samples, the probability of anomalies tends to be extremely low. Therefore, we define a probability-based anomaly measurement as a negative log conditional probability as follows:
\begin{align*}
 A_{prob}(t)=-\text{log}p(\textbf{x}_t|\textbf{x}_{t-\omega:t-1}).
\end{align*}
The higher this measurement of $\textbf{x}_{t}$ is, the more likely it is to be abnormal.

\subsubsection{Gradient-based Measurement} 
The score function gives the direction of change in the log probability and the magnitude of change in that direction. Even if the probability values of normal and abnormal samples are similar, a significant difference in the gradient allows us to distinguish an anomaly from normal behavior.
The gradient-based anomaly measurement defined by the norm of the conditional score function is given as
\begin{align*}
A_{grad}(t)=\left\|\nabla_{\textbf{x}_{t-\omega:t}}\text{log}p(\textbf{x}_{t-\omega:t}|\textbf{x}_{t-\omega:t-1})\right\|,
\end{align*} where both $\textit{l}_1$-norm and $\textit{l}_2$-norm can be used as the norm, $\|\cdot\|$.
Although both cases can be used in experiments, we observe that using $\textit{l}_1$-norm outperforms $\textit{l}_2$-norm, so we only adopt $\textit{l}_1$-norm in our experiments. The conditional score function can be replaced with conditional score network $S_{\boldsymbol{\theta}}(\textbf{x}_{t-\omega:t}^0, \textbf{x}_{t-\omega:t-1}^0, 0)$. 
We also describe the relationship between score function and log probability. By using the Taylor expansion and the Cauchy-Schwarz inequality, we can derive the following inequality: for any $\boldsymbol{\epsilon}>0$,
\begin{align*}
\left\|\nabla\text{log}p(\textbf{x})\right\|_2\geq \frac{|\text{log}p(\textbf{x}+\boldsymbol{\epsilon})-\text{log}p(\textbf{x})|} {\left\|\boldsymbol{\epsilon}\right\|_2}+O(\boldsymbol{\epsilon}).
\end{align*}
On a minimum point of the log-likelihood plane, the left term will be small, which means the log probability doesn't change in the neighborhood of $\textbf{x}$. To be more specific, the lower this measurement of $\textbf{x}_{t}$ is, the more likely it is to be normal. So we can consider the score function as the milestone of local minimum. Unlike other works which deal with only the probability, we focus on the local optimum, not on the point only, therefore we get a better strategy than other baselines. We demonstrate its performance in the experiments section.

\subsection{Calibrated Anomaly measurement Definitions with Purification}\label{Sec:purification}
However, the above three anomaly measurement definitions can be unstable when $\textbf{x}_{t-\omega:t-1}$ includes anomalous observations. For instance, $A_{recon}(t)$ should be high (resp. low) when $\textbf{x}_t$ is an anomalous (e.g., legitimate) observation, which is not always guaranteed in such a case (cf. the upper figure of Figure~\ref{fig:purification}). This phenomenon occurs for other two anomaly measurement definitions as well.



To alleviate this problem, we propose to purify, to get rid of anomalous observations if any, the condition part of the anomaly measurement definitions, i.e., $\textbf{x}_{t-\omega:t-1}$. We resort to the adversarial purification method (cf. Sec.~\ref{sec:ap}), i.e., adding slight noises to a sample via the forward process of SGM and denoising it via the reserve process. Throughout the denoising process, all non-legitimate signals can be removed from the sample.
Detailed purification and anomaly detection procedure is provided in Algorithm~\ref{algorithm}. Our purification step adopts the forward and reverse processes of SGMs to purify $\textbf{x}_{t-\omega: t-1}$ with the following steps (cf. Figure~\ref{fig:anomaly}): 

\begin{enumerate}
    \item This step is to blur the anomalies by adding noises to the conditional data. We first consider the concatenation of the conditional data $\textbf{x}_{t-\omega: t-1}$ and $0$, denoted by $\bar{\textbf{x}}_{t-\omega: t}$, to match its size with that of the input of $S_{\boldsymbol{\theta}}(\cdot,\cdot,\cdot)$. We perturb $\bar{\textbf{x}}_{t-\omega: t}$ from diffusion step $l=0$ to $l=\tau$ with forward SDE, where $\tau\in[0,1]$ is a hyperparameter to control the extent of noise added. $\bar{\textbf{x}}_{t-\omega: t}^{\tau}$ denotes the perturbed conditional data. (cf. Line 1-2 of Algorithm~\ref{algorithm}) 
    \item In this step, we produce a purified conditional data by gradually removing the noises (and potential anomalous values if any). We generate a sample $\bar{\textbf{x}}_{t-\omega: t}^0$ from $\bar{\textbf{x}}_{t-\omega: t}^{\tau}$ by solving the reverse SDE based on the trained conditional score network. We take the purified condition $\tilde{\textbf{x}}_{t-\omega: t-1}$ of length $\omega$ from the generated sample $\bar{\textbf{x}}_{t-\omega: t}^0$. 
    (cf. Line 3-5 of Algorithm~\ref{algorithm}).
\end{enumerate}

\begin{algorithm}[t]
\caption{Anomaly Detection Algorithm}
\label{algorithm}
\begin{flushleft}
\textbf{Input}: $\textbf{x}_{t-\omega:t-1}$ \\
\textbf{Parameter}: 
$\tau$ = A real parameter in $[0,1]$ to assign how much noise will be added to condition during purification. \\
\textbf{Output}: $\tilde{A}_{recon}$, $\tilde{A}_{prob}$, $\tilde{A}_{grad}$
\end{flushleft}
\begin{algorithmic}[1] 
\STATE ${\bar{\textbf{x}}}_{t-\omega:t}$ denotes concatenation of $\textbf{x}_{t-\omega:t-1}$ with zero vector.
\STATE Diffuse ${\bar{\textbf{x}}}_{t-\omega:t}$ until step $\tau$ with the forward SDE and thereby take ${\bar{\textbf{x}}}_{t-\omega:t}^{\tau}$.
\FOR{$l \in [0,\tau]$}
\STATE Run the sampling procedures to obtain $\tilde{\textbf{x}}_{t-\omega:t-1}$ from ${\bar{\textbf{x}}}_{t-\omega:t}^{\tau}$ with $S_{\boldsymbol\theta}(\bar{\textbf{x}}_{t-\omega:t}^{l},\textbf{0}, l)$.
\ENDFOR
\STATE Get $\textbf{z} \sim N(\textbf{0},\textbf{I}) $
\FOR{$l \in [0,1]$}
\STATE Run the sampling procedures to obtain  $\hat{\textbf{x}}_{t-\omega:t}$ from $\textbf{z}$ and \\ get $-\text{log}p(\textbf{x}_t|\tilde{\textbf{x}}_{t-\omega:t-1})$ \& $\left\|\nabla_{\textbf{x}_{t-\omega:t}}\text{log}p(\textbf{x}_{t-\omega:t}|\tilde{\textbf{x}}_{t-\omega:t-1})\right\|$ with $S_{\boldsymbol\theta}(\textbf{x}_{t-\omega:t}^{l},\tilde{\textbf{x}}_{t-\omega:t-1},l)$
\ENDFOR
\STATE \textbf{return} $\tilde{A}_{recon}$, $\tilde{A}_{prob}$, $\tilde{A}_{grad}$
\end{algorithmic}
\end{algorithm}

Since the model learns a distribution of the training data, which consist mostly of normal observations, we can restore to the most appropriate \emph{purified} sample from the potentially abnormal condition. Note that the difference will be larger (resp. small) if the condition contains anomalies (resp. normal observations) after the purification (cf. the lower figure of Figure~\ref{fig:purification}). By appropriately setting the hyperparameter $\tau$, we can control how aggressively the conditional data is purified. 

After purifying the condition, we achieve a calibrated condition, denoted $\tilde{\textbf{x}}_{t-\omega:t-1}$. By replacing the original condition with the purified one, we can get the following calibrated conditional score network, $S_{\boldsymbol{\theta}}(\textbf{x}_{t-\omega:t}^l, \tilde{\textbf{x}}_{t-\omega:t-1}, l) \approx \nabla_{\textbf{x}_{t-\omega:t}^l}{\text{log}p(\textbf{x}_{t-\omega:t}^l|\tilde{\textbf{x}}_{t-\omega:t-1})}$. We then generate a calibrated sample, $\tilde{\textbf{x}}_{t}$. Therefore we attain the following calibrated anomaly measurement definitions:
\begin{align*}
&\tilde{A}_{recon}(t)=\left\|\tilde{\textbf{x}}_t-{\textbf{x}}_t\right\|_2^2,\\
&\tilde{A}_{prob}(t)=-\text{log}p(\textbf{x}_t|\tilde{\textbf{x}}_{t-\omega:t-1}),\\
&\tilde{A}_{grad}(t)=\left\|\nabla_{\textbf{x}_{t-\omega:t}}\text{log}p(\textbf{x}_{t-\omega:t}|\tilde{\textbf{x}}_{t-\omega:t-1})\right\|.
\end{align*}

In order to achieve stable detection performance, the overall anomalous measurement $A_{anomaly}$ is calculated as the following seven cases by combining them: $\tilde{A}_{recon}$, $\tilde{A}_{prob}$, $\tilde{A}_{grad}$, $\tilde{A}_{recon}*\tilde{A}_{prob}$, $\tilde{A}_{recon}*\tilde{A}_{grad}$, $\tilde{A}_{prob}*\tilde{A}_{grad}$, and $\tilde{A}_{recon}*\tilde{A}_{prob}*\tilde{A}_{grad}$. It is worth mentioning that the above three measurements typically have different scales and their arithmetic mean is not appropriate. Therefore, we multiply them as in the term frequency-inverse document frequency (TF-IDF~\citep{rajaraman_ullman_2011}) widely used in natural language processing --- one can also consider that our definition is the geometric mean of the three measurements without the cube root. We also point out that multiplying two measurements has been used in previous anomaly detection work, TadGAN~\citep{geiger2020tadgan}, which mainly focuses on univariate time-series. We set the anomaly threshold as a hyperparameter and consider samples with anomaly measurements above the threshold to be anomalies.

\begin{table}[t]
\caption{Detailed characteristics of the benchmark datasets we use for our experiments}
\vspace{-0.2cm}
\centering
\begin{tabular}{c|c|c|c|c}
    \hline
    Dataset & \# of train set & \# of test set & Dim. & Ratio (\%) \\
    \hline
    SWaT & 496,800 & 449,919 & 51 & 12.14\\
    SMAP & 135,183 & 427,617 & 25 & 12.8\\
    MSL & 58,317 & 73,729 & 55 & 10.5 \\  
    PSM & 129,784 & 87,841 & 25 & 27.76\\ 
    SMD & 109,577 & 109,578 & 38 & 4.2\\
    \hline
\end{tabular}
\label{table10}
\vspace{-0.3cm}
\end{table}

\begin{table}[t]
\caption{The best hyperparameters of MadSGM in each dataset}
 \vspace{-0.2cm}
\centering
\resizebox{.95\columnwidth}{!}{
\begin{tabular}{c|c|c|c|c|c|c}
    \hline
    Dataset& Length & SDE type & $n_{layer}$ & $n_{resnet}$ & tol & $n_{iter}$ \\
    \hline
    SWaT& \multirow{5}{*}{10} & \multirow{5}{*}{VP} & 4 & 4 & $1e^{-3}$ & 210,000\\
    SMAP & &  & 3 & 2 & $1e^{-3}$ & 40,000\\
    MSL & & & 3 & 3  & $1e^{-3}$ & 100,000\\ 
    PSM  & &  & 3 & 2  & $1e^{-2}$ & 55,000 \\    
    SMD  & &  & 4 & 4  & $1e^{-3}$ & 45,000\\
    \hline
\end{tabular}
}
\vspace{-0.5cm}
\label{Table:hyperparameter}
\end{table}

\begin{table*}[t]
  \caption{   
    Experimental results in terms of $\text{F1}_{\text{PA}}$ and the area under the curve (AUC) of $\text{F1}_{\text{PA\%K}}$ with varying $K$. The best results are in boldface and the second ones are underlined. Avg.Rank denotes the average rank of each model.}
\vspace{-0.3cm}
 \setlength{\tabcolsep}{4pt}
\begin{center}  
   \begin{tabular}{c|cccccccccc|cc}
    \toprule
   \multirow{2}{*}{Method} & \multicolumn{2}{c}{SWaT} & \multicolumn{2}{c}{SMAP} & \multicolumn{2}{c}{MSL} & \multicolumn{2}{c}{PSM} & \multicolumn{2}{c}{SMD}  & \multicolumn{2}{|c}{Avg. Rank}
     \\ 
    \cmidrule(lr){2-3}\cmidrule(lr){4-5}\cmidrule(lr){6-7}\cmidrule(lr){8-9}\cmidrule(lr){10-11} \cmidrule(lr){12-13}
    & AUC & $\text{F1}_{\text{PA}}$   & AUC & $\text{F1}_{\text{PA}}$   &  AUC & $\text{F1}_{\text{PA}}$    & AUC & $\text{F1}_{\text{PA}}$   & AUC & $\text{F1}_{\text{PA}}$ & AUC & $\text{F1}_{\text{PA}}$  \\
    \hline
    OCSVM       &   0.2454  &   0.7926  &  0.3834  &   0.8126  &    0.286  &   0.6804  &   0.4711  &   0.6704  &   0.1199  &   0.2821  &   7.8  &   8.2  \\
DeepSVDD   &   0.7948  &   0.8802  &  0.3897  &   0.8235  &   0.3311  &   0.8218  &   0.6157  &   0.9133  &   0.1833  &   0.6981  &   4.4  &   5.4  \\
DAGMM   &   0.8021  &   0.8852  &  0.3491  &   0.8403  &   0.3266  &   0.7801  &   \underline{0.6244} &   0.9089  &   \underline{0.3317} &   0.8539  &   4.2  &   5.0  \\
LSTM-VAE    &   0.4412  &   0.6997  &  \underline{0.3989} &   0.9635  &   0.3465  &   0.8813  &   0.4557  &  0.5986  &   0.1264  &   0.4173  &   5.6  &   6.6  \\
OmniAnomaly &  0.2314  &  0.3398  &  0.3928  &   0.8461  &   0.2878  &   0.6797  &  0.4462  &   0.6161  &  0.1186  &  0.2667  &   8.2  &  8.4  \\
MAD-GAN     &   0.7782  &   0.8988  &  0.3672  &   0.8184  &   0.3349  &   0.8627  &   0.5754  &   0.9387  &   0.1647  &   0.6166  &   5.6  &   4.8  \\
TAnoGAN     &   0.8014  &   0.8486  &  0.3532  &   0.7862  &   \underline{0.3475} &   0.8371  &   0.6117  &   0.9643  &   0.2436  &   0.7493  &   4.0 &   5.6  \\
USAD        &   \underline{0.8197} &    0.868  &  0.3135  &  0.6958  &   0.3226  &  0.6254  &   0.5851  &    0.817  &   0.1972  &   0.4968  &   5.4  &   8.0  \\
AT          &   0.2328  &   \underline{0.9593} &  0.124  &   \underline{0.9667} &  0.2715  &   \textbf{0.9533} &   0.4799  &   \underline{0.9776} &   0.1204  &   \underline{0.9163} &  8.8  &   1.8 \\
\hline

Ours      &   \textbf{0.8273} &   \textbf{0.9651} &  \textbf{0.4075} &    \textbf{0.9690} &   \textbf{0.3609} &   \underline{0.9215} &   \textbf{0.6388} &     \textbf{0.9800} &   \textbf{0.3786} &   \textbf{0.9298} &   1.0  &   1.2 \\

    \bottomrule 
    \end{tabular}
    \vspace{-0.3cm}
    \label{Table:AD}
  \end{center}
\end{table*}

\section{Experiments}
In this section, we conduct experiments to illustrate the performance of the MadSGM on 
five real-world datasets from various fields with nine benchmark baselines. In particular, our collection of baselines covers various types of time-series anomaly detection methods, ranging from transformer-based models to VAEs and GANs. For the baselines, we reuse their released source codes in their official repositories and rely on their designed training procedures. Details of the software and hardware environment used in our experiments are as follow: \textsc{Ubuntu} 18.04 LTS, \textsc{Python} 3.9.12, \textsc{CUDA} 9.1, \textsc{NVIDIA} Driver 470.141, i9 CPU, and \textsc{GeForce RTX 2080 Ti}.

\subsection{Datasets}
We used five benchmark datasets for time-series anomaly detection in our experiments. The characteristics of datasets, including their data dimensions, the numbers of train and test samples, and anomaly ratios are summarized in Table~\ref{table10}. We briefly introduce them in the following:

\begin{itemize}
    \item \textit{Secure Water Treatment (SWaT)}~\citep{mathur2016swat}: The SWaT dataset was recorded over 11 days from a water treatment testbed, which has 26 sensor values and 25 actuator operations. 
    \item \textit{Mars Science
Laboratory rover (MSL)} and \textit{Soil Moisture Active Passive satellite (SMAP)}~\citep{hundman2018detecting}: Both MSL and SMAP are collected from spacecraft monitoring systems of NASA, which have 55 and 25 dimensions, respectively. Specifically, we used 53 out of 55 channels in the SMAP dataset excluding two P-2 channels.
    \item \textit{Pooled Server Metrics (PSM)}~\citep{abdulaal2021practical}: The PSM dataset is provided by eBay and consists of 25 features of sever machine metrics such as CPU utilization and memory collected internally from multiple application server nodes. 
    \item \textit{Server Machine Dataset (SMD)}~\citep{10.1145/3292500.3330672}: The SMD is server status log dataset collected from  28 different machines of a large internet company during 5 weeks. In this dataset, we use only four entities named as machine-1-1, 2-1, 3-2 and 3-7, respectively.
\end{itemize}

\subsection{Baselines}
We compare the MadSGM with several types of unsupervised anomaly detection methods:   density-based, boundary-based, and reconstruction-based methods. At first, DAGMM~\citep{zong2018deep} is used as the density-based method. Next, the boundary-based method contains OCSVM~\citep{tax2004support} with RBF kernel and DeepSVDD~\citep{ruff2018deep}. Both density-based and boundary-based models can't use a sliding window input since they are not designed to deal with the temporal dependency. Finally, we consider six reconstruction-based models for time-series anomaly detection, including VAE-based methods: LSTM-VAE~\citep{DBLP:journals/corr/abs-1711-00614} and OmniAnomaly~\citep{10.1145/3292500.3330672}; GAN-based methods: MAD-GAN~\citep{Li2019MADGANMA}, TAnoGAN~\citep{bashar2020tanogan}, and USAD~\citep{usad}; a Transformer-based method: AT~\citep{xu2022anomaly}.


\subsection{Evaluation Metrics}
Most works for the time-series domain have adopted the widely-used point adjustment approach, introduced by \citep{xu2018unsupervised}: if any time point in a successive anomaly segment is detected, all observations in this segment are regarded to be correctly detected as anomalies. The F1-score with the point-adjust way denoted as $\text{F1}_{\text{PA}}$ is more suitable for range-based anomalies than the naive F1-score (F1). The $\text{F1}_{\text{PA}}$ will be higher than the F1. 

However, \citet{kim2022towards} argued some limitations in which $\text{F1}_{\text{PA}}$ has a high possibility to be overestimated. They provided empirical evidence that a random anomaly measurement outperformed  state-of-the-art methods on almost all benchmark datasets. For this reason, they proposed an alternative evaluation metric named by $\text{PA\%K}$ which can remedy both the overestimation of $\text{F1}_{\text{PA}}$ and underestimation of $\text{F1}$. 

Let us define $S_m=:\{t_s^m, \ldots, t_e^m\}$ as an anomaly segment for $m = 1, \ldots, M$ and $t_s^m$ and $t_e^m$ are the  start and end times of $S_m$, respectively. The $\text{PA\%K}$ protocol is defined as follows:
 \begin{equation*}
     \hat{y}_t = \begin{cases}
  1  & \text{ if} \,\, A(\textbf{x}_{t-\omega:t}) > \delta \,\, \text{ or}\\
    & t \in S_m \text{ and } \dfrac{ | \{t'\,|\, t' \in S_m, A(\textbf{x}_{t^{'}-\omega:t^{'}}) > \delta \}| }{|S_m|} > K \\
  0 &  \text{ otherwise}
\end{cases},
 \end{equation*} 
 where $\hat{y}_t$ is a predicted label and $\delta$ is a certain threshold, $A(\cdot)$ is the anomaly measurement of input, $|\cdot|$ is the cardinality of a set, and $K \in [0, 1]$ is a ratio.
We denote $\text{F1}_{\text{PA\%K}}$ as F1-score with the $\text{PA\%K}$ strategy.

\begin{figure*}[t]
     \centering
     \begin{subfigure}[b]{0.3\textwidth}
         \centering
         \includegraphics[width=\textwidth]{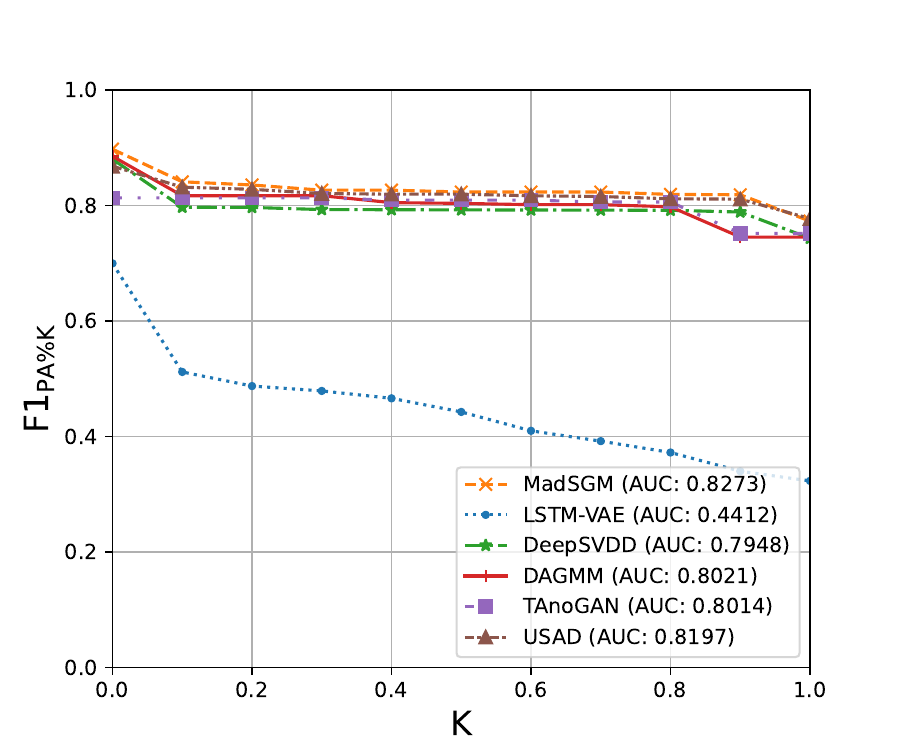}
         \caption{SWaT}
         \label{fig3:swat}
     \end{subfigure}
     \hfill
     \begin{subfigure}[b]{0.3\textwidth}
         \centering
         \includegraphics[width=\textwidth]{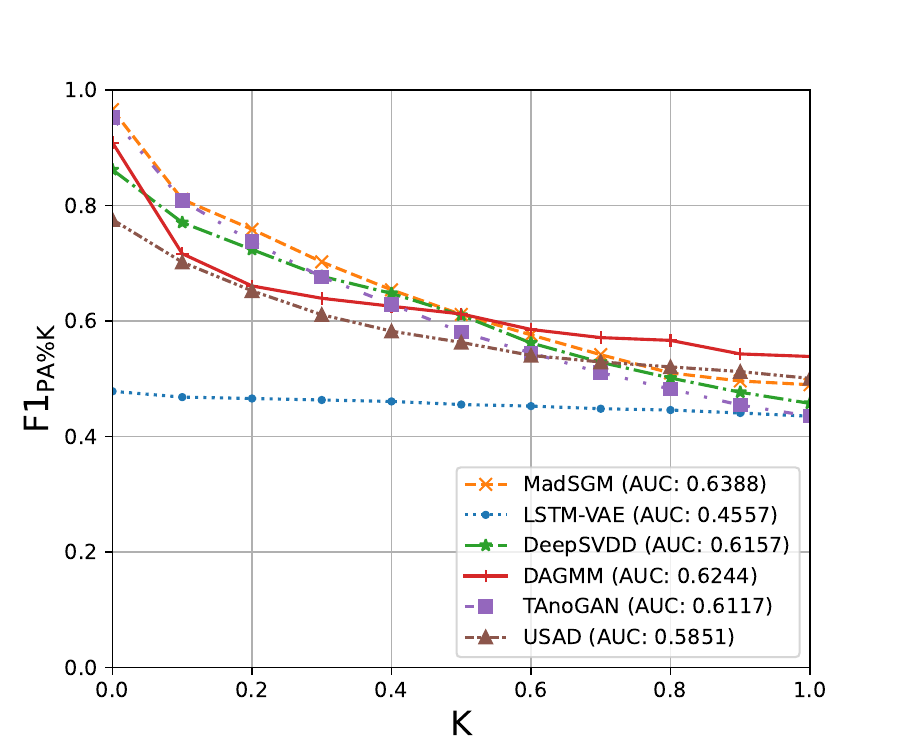}
         \caption{PSM}
         \label{fig3:psm}
     \end{subfigure}
     \hfill
     \begin{subfigure}[b]{0.3\textwidth}
         \centering
         \includegraphics[width=\textwidth]{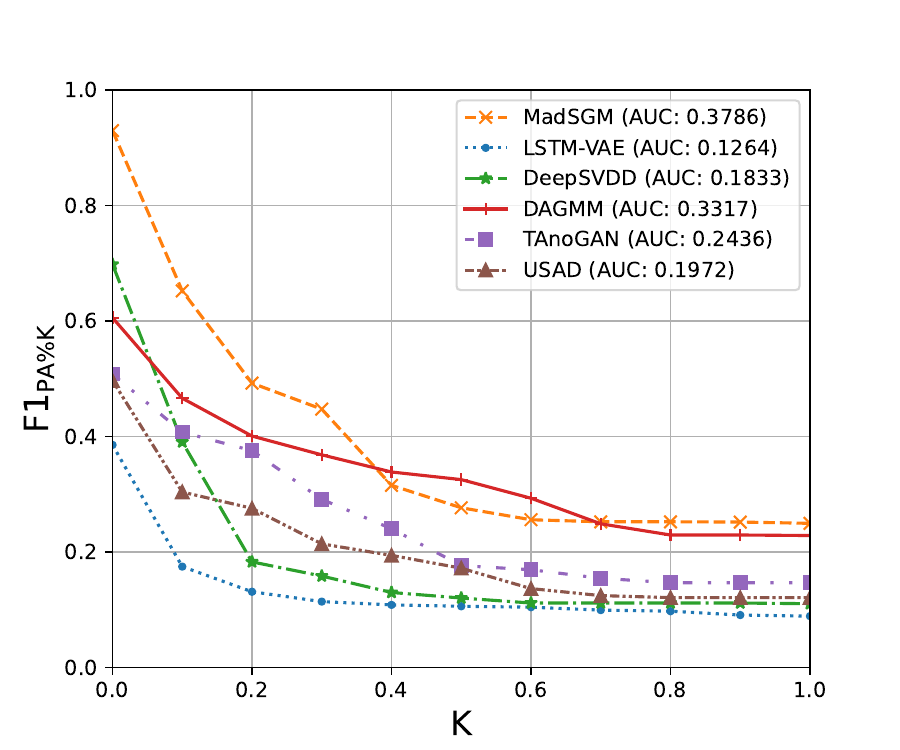}
         \caption{SMD}
         \label{fi3:smd}
     \end{subfigure}
        \vspace{-0.1cm}
        \caption{The curve of $\text{F1}_{\text{PA\%K}}$ for SWaT, PSM, and SMD datasets by varying $\text{K}$ from 0 to 1. AUC denotes the area under the curve. The higher the AUC value, the better the detection performance.}        
        \label{fig3}
\vspace{-0.1cm}
\end{figure*}

In this paper, we used the area under the curve (AUC) of $\text{F1}_{\text{PA\%K}}$ as the main evaluation metric obtained by increasing $K$ from 0 to 1 by 0.1. Here, $\text{F1}_{\text{PA\%K}}$ with $K = 0$ and $K = 1$ is identical with the $\text{F1}_{\text{PA}}$ and $\text{F1}$, respectively.
\autoref{fig3} shows that $\text{F1}_{\text{PA\%K}}$ values with different $K$ and AUCs of the proposed model and some baselines on benchmark datasets.

\subsection{Hyperparameters}\label{Sec:hyperparameters}
 
We describe the best hyperparameters of our proposed model for reproducibility. Table~\ref{Table:hyperparameter} provides our best hyperparameters. We set the length (window size) of time-series sequences to 10 and use VP as the type of SDE for all datasets. In the architecture of the conditional score network, $n_{layer}$ and $n_{resnet}$ mean the depth of U-net and the number of residual blocks in each layer, respectively. We search $n_{layer}$ in $\{3,4\}$ and $n_{resnet}$ in $\{2,3,4\}$. When we calculate the exact likelihood, we consider $\{1e^{-2}, 1e^{-3}\}$ as the tolerance level of probability flow ODE. For training iterations $n_{iter}$, the optimal setting depends on the dataset and we check it every 5000 iterations. For other hyperparameter settings in SGMs, we follow that of VP SDE in \citet{song2021SDE}.



\begin{figure}[H]
\centering
\vspace{-0.3cm}
\includegraphics[width=0.47\textwidth]{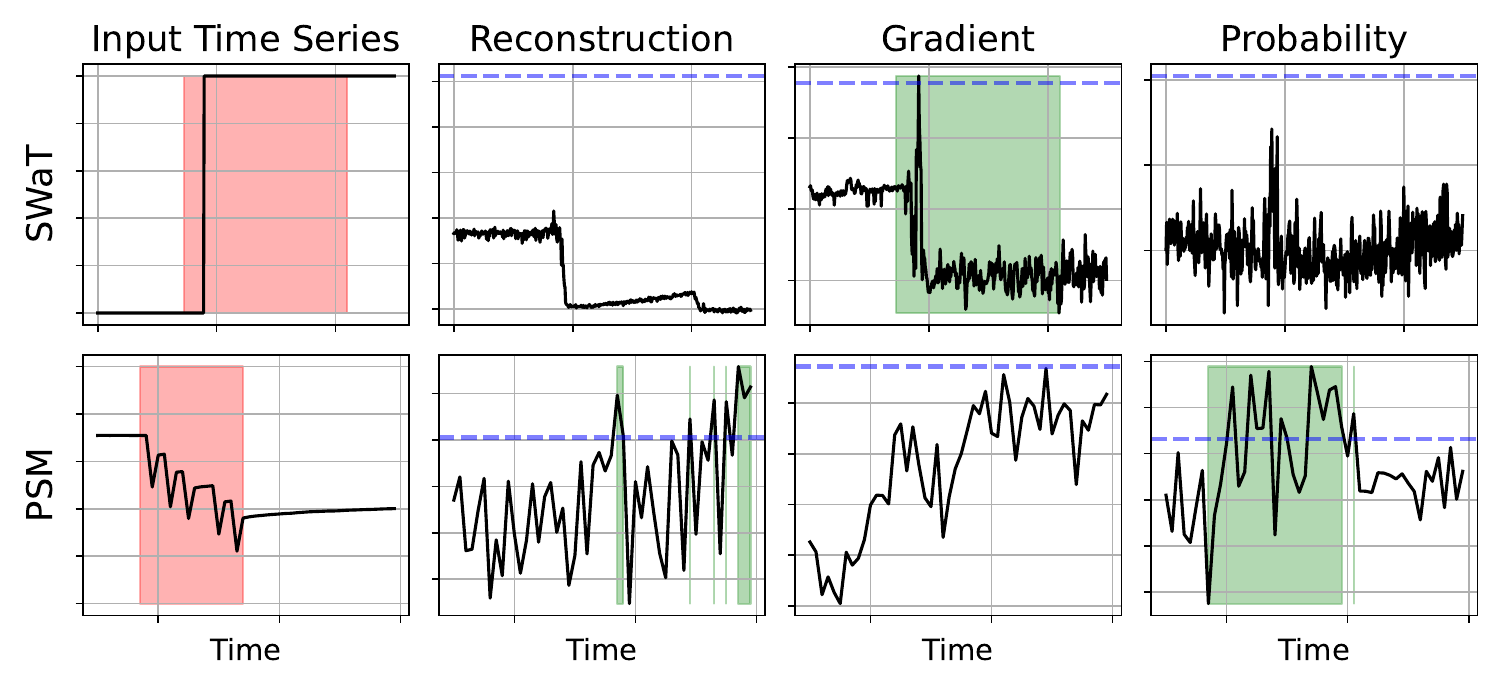} 
\vspace{-0.3cm}
\caption{Examples of anomalies that are not detectable by one or two of the three anomaly measurements but by another. Red and green indicate true anomalies and  estimated anomalies based on anomaly measurements, respectively. Blue dashed lines represent thresholds of anomaly measurements to detect anomalous observations.}
\label{fig:anomaly_ex}
\vspace{-0.5cm}
\end{figure}

\subsection{Experimental Results}
We compare our proposed methods with several popular anomaly detection models on five real-world datasets. We calculate evaluation metrics after a certain time point since time-series anomaly detectors use different sizes of the sliding window. Note that MadSGM generates definite sample since it uses probability flow ODE, which follows fixed path (see Section~\ref{Sec:trainingsampling}). Therefore we achieve constant results on various seeds, like other non-generative models.

Table~\ref{Table:AD} provides that the MadSGM performs well in all datasets with the highest or the second-best results. 
Especially, for all datasets, the MadSGM shows overwhelming performance in terms of the AUC of $\text{F1}_{\text{PA\%K}}$. For $\text{F1}_{\text{PA}}$, the MadSGM outperforms all the baselines on all datasets except MSL.  \autoref{fig3} demonstrates that for all $K$, the proposed methods has higher values of $\text{F1}_{\text{PA\%K}}$ than LSTM-VAE, DeepSVDD, and TAnoGAN. By the average rank of the last column in Table~\ref{Table:AD}, it generally has a more reliable performance than the current state-of-the-art irrespective of evaluation metrics, which shows its robust detection performance. In other words, the proposed method can cope with all datasets using the broadest ever set of anomaly measurements, while most existing methods work well only with specific datasets. For instance, whereas DAGMM earns the second-best results on the PSM and SMD datasets, for the SWaT and SMAP datasets with long-time sequences, it doesn't achieve reasonable performance.

Furthermore, when simultaneously considering both metrics, $\text{F1}_{\text{PA}}$ and AUC, the excellence of our proposed method is also demonstrated. When checking only one metric, either $\text{F1}_{\text{PA}}$ or AUC, baselines sometimes show good performance. However, when checking both the metrics, it is observed that although the result of one metric is reasonable, that of the other is poor. For example, DeepSVDD and DAGMM have decent performance for the AUC of $\text{F1}_{\text{PA\%K}}$, but not for $\text{F1}_{\text{PA}}$. In addition, AT and MAD-GAN of time-series anomaly detectors are somewhat overestimated from the PA\%K protocol's perspective, because there are discrepancies in rankings between the $\text{F1}_{\text{PA}}$ and AUC of $\text{F1}_{\text{PA\%K}}$. However, in all cases, the performance of MadSGM is the best or the second-best for both metrics. Therefore, the effectiveness of our proposed method is demonstrated by the fact that MadSGM has an overwhelming performance in Table~\ref{Table:AD}, regardless of datasets and metrics.

\begin{table*}[t]
  \caption{Experimental results according to various anomaly measurement settings in the proposed method.}
\vspace{-0.3cm}
 \setlength{\tabcolsep}{4pt}
\begin{center}  
    { 
   \begin{tabular}{c|cccccccccc|cc}
    \toprule
   \multirow{2}{*}{Anomaly measurement} & \multicolumn{2}{c}{SWaT} & \multicolumn{2}{c}{SMAP} & \multicolumn{2}{c}{MSL} & \multicolumn{2}{c}{PSM} & \multicolumn{2}{c}{SMD}  & \multicolumn{2}{|c}{Avg. Rank}
     \\ 
    \cmidrule(lr){2-3}\cmidrule(lr){4-5}\cmidrule(lr){6-7}\cmidrule(lr){8-9}\cmidrule(lr){10-11} \cmidrule(lr){12-13}
    & AUC & $\text{F1}_{\text{PA}}$   & AUC & $\text{F1}_{\text{PA}}$   &  AUC & $\text{F1}_{\text{PA}}$    & AUC & $\text{F1}_{\text{PA}}$   & AUC & $\text{F1}_{\text{PA}}$  & AUC & $\text{F1}_{\text{PA}}$ \\
    \hline
    $A_{recon}$ & 0.8130 & 0.9327 &   0.3643 & 0.9639 &   0.3335 & 0.7633 &   0.6074 & 0.9795 &   0.2609 & 0.8763 &       5.0 &     4.0 \\
    $A_{prob}$ & 0.7927 & \textbf{0.9651} &   0.3314 & 0.8461 &   \textbf{0.3609} & \textbf{0.9215} &   0.6337 & 0.9660 &   \textbf{0.3786} & \textbf{0.9298} &       3.4 &     3.2 \\
    $A_{grad}$ & 0.4358 & 0.9010 &   \textbf{0.4075} & 0.9650 &   0.3511 & 0.8647 &   \underline{0.6351} & \textbf{0.9800} &   0.2505 & 0.8843 &       4.0 &     3.4\\
    $A_{recon}*A_{prob}$ & \underline{0.8257} & 0.9336 &   0.3932 & \underline{0.9665} &   0.3333 & 0.7450 &   0.6162 & 0.9724 &   0.2638 & 0.8679 &       4.2 &     4.8 \\
    $A_{recon}*A_{grad}$ & 0.8123 & 0.9358 &   0.3642 & 0.9648 &   0.3360 & 0.7756 &   0.6093 & \underline{0.9799} &   0.2544 & 0.8701 &       5.2 &     3.8 \\
    $A_{prob}*A_{grad}$ &  0.7918 & \underline{0.9510} &   0.3337 & 0.9022 &   \underline{0.3542} & \underline{0.8979} &   \textbf{0.6388} & 0.9660 &   \underline{0.3648} & \underline{0.9169} &       3.4 &     3.8 \\
    $A_{recon}*A_{prob}*A_{grad}$ & \textbf{0.8273} & 0.9318 &   \underline{0.3945} & \textbf{0.9690} &   0.3412 & 0.7645 &   0.6181 & 0.9722 &   0.2640 & 0.8707 &       2.8 &     4.4 \\
    \bottomrule 
    \end{tabular}
  }
    \label{Table:variousAD}
    \vspace{-0.2cm} 
  \end{center}
\end{table*}

\begin{table*}[t]
  \caption{Experimental results by changing $\tau$, a hyperparameter for the diffusion step of the forward SDE in the purification process.}
\vspace{-0.4cm}
 \setlength{\tabcolsep}{4pt}
\begin{center}  
    { 
   \begin{tabular}{c|cccccccccc|cc}
    \toprule
    \multirow{2}{*}{$\tau$}& \multicolumn{2}{c}{SWaT} & \multicolumn{2}{c}{SMAP} & \multicolumn{2}{c}{MSL} & \multicolumn{2}{c}{PSM} & \multicolumn{2}{c}{SMD}  & \multicolumn{2}{|c}{Avg. Rank} 
     \\ 
    \cmidrule(lr){2-3}\cmidrule(lr){4-5}\cmidrule(lr){6-7}\cmidrule(lr){8-9}\cmidrule(lr){10-11} \cmidrule(lr){12-13}
     & AUC & $\text{F1}_{\text{PA}}$   & AUC & $\text{F1}_{\text{PA}}$   &  AUC & $\text{F1}_{\text{PA}}$    & AUC & $\text{F1}_{\text{PA}}$   & AUC & $\text{F1}_{\text{PA}}$  & AUC & $\text{F1}_{\text{PA}}$  \\
    \hline
    0.0  &   0.8040 & \textbf{0.9651} &   0.3377 & 0.9399 &   0.3441 & \underline{0.8979} &   \textbf{0.6388} & \underline{0.9799} &   \textbf{0.3786} & \textbf{0.9298} &       4.0 &     2.4 \\
    0.05 &   0.8145 & \underline{0.9358} &   \underline{0.3945} & 0.9650 &   0.3476 & 0.8780 &   0.6092 & 0.9616 &   0.1954 & 0.8115 &       4.4 &     4.2 \\
    0.1  &   0.8217 & 0.9091 &   \textbf{0.4075} & 0.9665 &   0.3487 & \textbf{0.9215} &   0.6033 & 0.9686 &   0.2609 & \underline{0.8843} &       3.8 &     2.8 \\
    0.15 &   0.8175 & 0.9081 &   0.3904 & \underline{0.9686} &   \textbf{0.3609} & 0.8740 &   0.6076 & 0.9726 &   0.2638 & 0.8548 &       3.4 &     3.6 \\
    0.2  &   \underline{0.8232} & 0.8994 &   0.3730 & \textbf{0.9690} &   \underline{0.3542} & 0.8711 &   \underline{0.6226} & 0.9691 &   \underline{0.3362} & 0.8575 &       2.6 &     4.2 \\
    0.25 &   \textbf{0.8273} & 0.9014 &   0.3711 & 0.9648 &   0.3495 & 0.8734 &   0.6304 & \textbf{0.9800} &   0.3210 & 0.8822 &       2.8 &     3.8 \\
    \bottomrule 
    \end{tabular}
  }
    \label{Table:purification}
    \vspace{-0.3cm} 
  \end{center}
\end{table*}
\begin{table}[t]
  \caption{Experimental results by changing a solver method for the probability flow ODE.}\label{tbl:solver}
\vspace{-0.3cm}
 \setlength{\tabcolsep}{4pt}
\begin{center}  
    {
   \begin{tabular}{ccccc}
    \toprule
   \multirow{2}{*}{Solver}& \multicolumn{2}{c}{MSL} & \multicolumn{2}{c}{PSM} \\ 
   \cmidrule(lr){2-3}\cmidrule(lr){4-5}
    & AUC & $\text{F1}_{\text{PA}}$ & AUC & $\text{F1}_{\text{PA}}$
     \\ 
    \hline
    RK45 & 0.3609 & 0.9215 & 0.6388 & 0.9800 \\
    RK23 & 0.3610 & 0.9075 & 0.6252 & 0.9804 \\
    DOP853 & 0.3598 & 0.9010 & 0.6256 & 0.9803 \\
    \bottomrule 
    \end{tabular}
  }
    \label{Table:sensitivity}
    \vspace{-0.5cm} 
  \end{center}
\end{table}

\section{Ablation Study}

\subsection{Ablation study on Anomaly measurements}\label{sec:abl_anomaly_score}
In this section, we conduct anomaly detection tasks by varying the type of anomaly measurements. As in Section~\ref{Sec:anomalyscore}, our method evaluates 3 types of anomaly measurements: i) reconstruction-based measurement, $A_{recon}$, ii) probability-based measurement, $A_{prob}$, and iii) gradient-based measurement, $A_{grad}$. When we consider all combinations of 3 types, by multiplying each other, there are 7 types of anomaly measurements in total. Table~\ref{Table:variousAD} shows that the best type of anomaly measurement differs depending on datasets and evaluation metrics. In other words, one type of anomaly measurement doesn't always give the best performance. We also give representative visualizations to support our claims in Figure~\ref{fig:anomaly_ex}. Therefore, this subsection demonstrates the importance of considering various types of anomaly measurements simultaneously to achieve the best performance in anomaly detection --- note that our proposed method considers a broader set of anomaly measurements than any other baseline methods (see Table~\ref{tab:anoscore}). 




\subsection{Ablation study on Purification}
We summarize the experimental results by changing the hyperparameter $\tau$, which refers to the diffusion step of the forward SDE in the purification process (cf. Section~\ref{Sec:purification}). We consider the scale of $\tau$ in $\{0, 0.05, 0.1, 0.15, 0.2, 0.25\}$. Especially, $\tau=0$ means to perform anomaly detection without purification. In this section, the reason why the purification process is needed is confirmed. In Table~\ref{Table:purification}, conducting anomaly detection without the purification (i.e., $\tau=0$) achieves reasonable results. However, for some datasets and evaluation metrics, the purification process (i.e., $\tau>0$) is helpful for improving performance. In particular, the performance of SMAP and MSL was significantly improved by using the purification strategy. These meaningful results come from the effect of blurring and purifying anomalies in the conditional data which makes a significant difference between the observed and reconstructed (generated) values (see Figure~\ref{fig:purification}). In this regard, Table~\ref{Table:purification} shows the efficacy of our purification process.

\subsection{Ablation study on a solver method for the probability flow ODE}
By changing a solver method for the probability flow ODE in Section~\ref{Sec:trainingsampling}, we observe the effect of the solver on the anomaly detection task. We test a total of 3 solvers: explicit Runge-Kutta methods with order 5(4) (RK45, default)~\citep{rk45,rk452}, order 3(2) (RK23)~\citep{rk23}, and order 8 (DOP853)~\citep{dop}. Table~\ref{tbl:solver} shows the result of this ablation study in MSL and PSM. In almost all cases, there are only negligible differences in detection results among the three solvers. However, RK23 and DOP853 often have a large drop in their performance, compared to RK45. For example, for AUC in PSM, the results of RK23 and DOP853 are quite lower than that of RK45. Therefore, we choose RK45 as the main solver for the probability flow in other experiments.

\section{Conclusion}
Time-series anomaly detection is a long-standing research topic in the field of machine learning. In this work, we proposed MadSGM, a novel framework based on SGMs for anomaly detection by designing a conditional score network and its denoising score matching loss. Especially, our method is distinguished from other existing methods for anomaly detection in that i) it can employ the most comprehensive set of anomaly measurements and ii) there exists the purification process to eradicate potentially misleading (conditional) inputs. The first strategy enables capturing diverse anomaly patterns from time series data with complicated characteristics, and the other helps enlarge the difference between the observed and reconstructed observation and thereby, it becomes easier to detect anomalies with our method even when the input has noises. Our extensive experiments with five benchmark datasets and nine baselines successfully demonstrate that these two strategies make the performance of anomaly detection improve severely, showing that MadSGM provided outstanding performance compared to other anomaly detection methods.

\section*{Acknowledgement}
Noseong Park is the corresponding author. This work was supported by the Institute of Information \& Communications Technology Planning \& Evaluation (IITP) grant funded by the Korean government (MSIT) (No.2020-0-01361, Artificial Intelligence Graduate School Program (Yonsei University)).

\clearpage
\bibliographystyle{ACM-Reference-Format}
\balance
\bibliography{example_paper}

\end{document}